\documentclass{article}

\usepackage{neurips_2021}

\usepackage[utf8]{inputenc} 
\usepackage[T1]{fontenc}    
\usepackage{hyperref}       
\usepackage{url}            
\usepackage{booktabs}       
\usepackage{amsfonts}       
\usepackage{nicefrac}       
\usepackage{microtype}      
\usepackage{xcolor}         
\usepackage{amsmath}
\usepackage{graphicx}
\usepackage{wrapfig}
\usepackage[normalem]{ulem}

\makeatletter

\newcommand{\thickhline}{
\noalign {\ifnum 0=`}\fi \hrule height 1.0pt
\futurelet \reserved@a \@xhline
}
\usepackage{amsmath}
\DeclareMathOperator*{\argmax}{\arg\!\max}
\makeatother

\title{Pay Better Attention to Attention: Head Selection in Multilingual and Multi-Domain Sequence Modeling}

\author{
    Hongyu Gong, Yun Tang, Juan Pino, Xian Li \\
    Facebook AI Research \\
    \texttt{\{hygong, yuntang, juancarabina, xianl\}@fb.com}
}

\begin{document}

\maketitle

\begin{abstract}
Multi-head attention has each of the attention heads collect salient information from different parts of an input sequence, making it a powerful mechanism for sequence modeling.
Multilingual and multi-domain learning are common scenarios for sequence modeling, where the key challenge is to maximize positive transfer and mitigate negative transfer across languages and domains.
In this paper, we find that non-selective attention sharing is sub-optimal for achieving good generalization across all languages and domains. We further propose attention sharing strategies to facilitate parameter sharing and specialization in multilingual and multi-domain sequence modeling.
Our approach automatically learns shared and specialized attention heads for different languages and domains to mitigate their interference. 
Evaluated in various tasks including speech recognition, text-to-text and speech-to-text translation, the proposed attention sharing strategies consistently bring gains to sequence models built upon multi-head attention. For speech-to-text translation, our approach yields an average of $+2.0$ BLEU over $13$ language directions in multilingual setting and $+2.0$ BLEU over $3$ domains in multi-domain setting.
\end{abstract}

\section{Introduction}

Recent progress on deep learning models, in particular multi-head attention, has brought significant gains to sequence modeling tasks including speech recognition \citep{moritz2020streaming}, text-to-text translation \citep{DBLP:conf/nips/VaswaniSPUJGKP17},  and speech-to-text translation \citep{DBLP:conf/iberspeech/VilaEFC18,DBLP:conf/interspeech/GangiNT19}.
Attention mechanism allows a model to focus on informative parts of the inputs, and multi-head attention computes attention over inputs by multiple heads independently. With each head attending to different information, multi-head attention potentially captures more complicated data patterns and extracts sophisticated knowledge.

Sequence modeling has attracted a lot of research interest in multilingual and multi-domain settings, where a model is trained on data in multiple language directions and data from different domains respectively. Key advantages of these settings are better data efficiency and the support of knowledge transfer among languages or domains. This is critical for resource-limited scenarios. For example, multilingual translation enhances the performance of low-resource languages via knowledge transfer from high-resource languages \citep{gu2018universal,DBLP:conf/asru/InagumaDKW19}. Given the data scarcity in individual domains, a common practice is to combine the data from various domains to augment the training set \citep{wang2020go}. Another appealing aspect of multilingual or multi-domain models is their low deployment and maintenance costs compared with numerous models trained for individual language pairs or domains.

Despite the positive knowledge transfer, negative interference has also been observed in multilingual (or multi-domain) training especially when languages (or domains) are dissimilar. Recent studies reveal from the optimization perspective that conflicting gradients in shared parameters is one cause of interference between languages (or domains) \citep{yu2020gradient}. A promising direction for interference mitigation is to design better strategies of parameter sharing. In some previous works, sharing is based on the similarity between languages (or domains), which require expert knowledge or pre-computed relatedness \citep{wu2019understanding}. Recent studies also propose branches and components specific to languages (or domains) in addition to shared modules \citep{bapna2019simple,guo2020learning}. 

In this work, we bring the mitigation of language and domain interference under a common umbrella, and tackle it by improving parameter sharing within multi-head attention. We propose strategies to select attention heads for different languages or domains. Instead of sharing everything across languages or domains, our model automatically learns to share heads among a subset of languages or domains. It encourages positive transfer within the subset and preserves their specificity without interference from outside the subset.
The major contributions of this work are summarized below: 

1. We propose attention head selection to mitigate interference in multilingual and multi-domain modeling;

2. The parameter sharing strategies are lightweight and preserve computational efficiency;

3. We extensively evaluate attention sharing strategies on various sequence modeling tasks including speech recognition, text-to-text and speech-to-text translation. Consistent gains are achieved 
across multiple benchmark datasets.

The paper is structured as follows. Section \ref{sec:related} discusses related works on sequence modeling in multilingual and multi-domain setting. In Section \ref{sec:model}, we introduce the proposed strategies of head selection in multi-head attention. Section \ref{sec:experiment} describes the empirical evaluation, followed by a discussion in Section \ref{sec:discussion}. We conclude this paper in Section \ref{sec:conclusion}.

\section{Related Work}
\label{sec:related}

\textbf{Multilingual learning}. 
Multilingual modeling has the potential to improve low-resource language performance through knowledge transfer from high-resource languages,
and it draws great interest from researchers in speech recognition and translation~\citep{Pratap2020MassivelyMA,Heigold2013MultilingualAM,Johnson2017GooglesMN,Dabre2020ASO,liu2020multilingual,Inaguma2019MultilingualES,Li2020MultilingualST}.  
Although impressive progress has been made for low-resource or zero-shot tasks, it is also found the multilingual model has inferior performance on high-resource tasks due to multilingual interference. 
In order to address this issue, some works focus on multilingual models with task-specific parameters. 
Different parameter sharing strategies are examined for the Transformer model \citep{Sachan2018ParameterSM}. 
Attention dependent on target languages is proposed to enhance the multilingual translation performance \citep{Blackwood2018MultilingualNM}. 
Treating multilingual modeling as an adaptation problem, \citet{bapna2019simple} first build a general multilingual model for all languages and then finetune newly added residual adapters for each language pair.
Another thread of work is to increase the model capacity to compensate for the high-resource language loss \citep{Pratap2020MassivelyMA}. 
~\citet{Shazeer2017OutrageouslyLN} propose mixture-of-experts and select RNN cells based on input tokens. ~\citet{Lepikhin2020GShardSG} extend it to Transformer with FFN experts. 
Different from previous works, we propose strategies of attention sharing among languages in the level of attention heads for multilingual modeling. 

\textbf{Multi-domain learning}. Similar to multilingual learning, multi-domain learning (MDL) can effectively utilize data from different domains but also suffers from interference due to inter-domain heterogeneity~\citep{Saunders2021DomainAA,pham2021revisiting}. 
Previous works address this issue from two perspectives: optimization and model architecture. For the optimization aspect, attempts have been made to synchronize the learning speed of different tasks~\citep{Chen2018GradNormGN}, adjust the gradients of individual tasks to alleviate gradient conflicts~\citep{yu2020gradient} and apply regularization to achieve better generalization in different domains~\citep{Dakwale2017FineTuningFN,Khayrallah2018RegularizedTO,Thompson2019OvercomingCF}. In terms of model architecture, domain-specific labels~\citep{Kobus2017DomainCF}, word embedding~\citep{Zeng2018MultiDomainNM}, sub-networks~\citep{wang2020go} are adopted to address the issue of domain divergence. The architecture can be specified during the general training with the mixed data from multiple domains~\citep{wang2020go} or during the finetuning in individual domains~\citep{bapna2019simple}. In this work, we deal with domain interference by leveraging domain-specific attention heads in multi-head attention.


\textbf{Attention selection}.  Selective self-attention networks proposes to apply masking to the inputs and pay more attention to content words \citep{Geng2020HowDS}. 
~\citet{Liu2021GumbelAttentionFM} selects text-related image regions with attention in multi-modality translation.
Compared to their methods,  we conduct automatic attention head selection for different tasks and focus on mitigating task interference.







\section{Model}
\label{sec:model}

In this section, we start with preliminaries of multi-head attention, and introduce our approach to attention interference mitigation. We put multilingual and multi-domain sequence modeling under the same umbrella in this study. For the simplicity of the following discussions, we refer to the two settings as multi-task modeling, where a task is one language or one domain.
Different from the standard multi-head attention, our model provides more attention heads than those used in computation. Different subsets of heads are assigned to each task so that partial attention sharing enables knowledge transfer and meanwhile mitigates interference. We introduce latent variables to modulate head selection, and propose strategies to learn the head assignment to different tasks.




\subsection{Preliminary}

\textbf{Multi-head attention}. As a core module in Transformer, multi-head attention paramterizes each head  with key, query and value transformation matrices \citep{DBLP:conf/nips/VaswaniSPUJGKP17}. The token representation is transformed into key, query and value vectors via these transformations. Each head assigns the attention of this token over the input sequence based on the matching between its query vector and key vectors of other tokens. The value vectors are weighted by the attention as the contextualized token representation. It is passed through linear projection as the output of the attention head. Suppose that head $h$ has output $\mathbf{x}^{(h)}$. Multi-head attention with $H$ heads yields an output $\mathbf{x}$ for the given token, which is the concatenation of all head outputs.
\begin{align}
    \mathbf{x} = \mathbf{x}^{(1)} \oplus \cdots \oplus \mathbf{x}^{(h)} \oplus \cdots \oplus \mathbf{x}^{(H)},
\end{align}
where $\oplus$ is vector concatenation. 


\textbf{Interference}. Maximal parameter sharing aims to learn universal knowledge across languages \citep{wang2020negative} and domains \citep{zeng2018multi}. To capture the task specificity, different languages or domains compete for model capacity, which is observed as the interference in previous studies. The interference results in degraded performance in jointly trained models. However, few works look into the improvement of parameter sharing within multi-head attention. This study explores head selection strategies to mitigate the inference in multilingual and multi-domain models.

\subsection{Latent Variable for Head Selection}



First, we outline our approach to learn a more general-purpose multi-head attention in Transformer from the Bayesian neural network perspective. Suppose that the input sequence is $x$ and the output sequence is $y$.
For conditional sequence modeling tasks such as machine translation, the posterior of $p(y\mid x)$ can be computed by marginalizing over the posterior of latent variable $z$, which modulates parameters ${\Theta}$ in the standard Transformer architecture:
\begin{equation}
\begin{split}
\label{eq:loglik}
     p(y\mid x, \Theta) = \mathbf{E}_{p(z\mid\Theta)}[p(y\mid x, z)] = \int p(y\mid x, z) p(z|\Theta)\, \mathrm{d}z 
\end{split}
\end{equation}

\paragraph{Parameterization of $z$.} In this work, we define $z$ as modulating the selection of attention heads. 
We model $z_{t}^{(h)}$ as a discrete latent variable from a Bernoulli distribution with  $ z_{t}^{(h)} \sim \mathcal{B}(\pi) \text{, } \pi \in [0, 1]$ indicating whether task $t$ selects attention head $h$.
This modeling choice allows us to prune attention heads during computation, which preserves computation efficiency as well as regularizes training.

Marginalizing over $z$ is intractable given numerous heads in neural models.
Therefore, we use variational inference to derive an approximate solution. Specifically, we learn an inference network $q_{\phi}(z)$, which is paramterized with $\phi$, to approximate the true distribution $p(z)$ and optimize the evidence lower bound (ELBO) of Eq. \ref{eq:loglik}: 
\begin{equation}
   \label{eq:lb}
   \log p(y \mid x) \geq \mathbf{E}_{q_{\phi}(z)} [\log p_\theta(y\mid x, z)] 
    - \text{KL}(q_{\phi}(z) \parallel p(z)), 
\end{equation}
where $\text{KL}$ is the KL-divergence between two distributions. In our work, we assume identical probability of each head being selected. Therefore, we have $p(z=1)=\frac{H}{H'}$, where $H$ and $H'$ are numbers of selected heads and all head candidates.



\paragraph{Training and interference.} We use the Gumbel-Softmax reparameterization \citep{jang2016categorical} to draw samples of $z$ from the posterior $q_{\phi}(z)$. It makes the model end-to-end differentiable, while learning discrete policies of head selection without resorting to policy gradients. We adopt a lightweight estimator of $q_{\phi}(z)$ by directly learning the logit parameters $\{\phi_{t}^{(h)}\}$: 
\begin{align}
     q_{\phi}(z_{t}^{(h)})= \frac{\exp((\phi_{t}^{(h)}(1) + \epsilon(1))/\tau)}{\sum_{j\in\{0,1\}} \exp((\phi^{(h)}(j) + \epsilon(j))/\tau)} \text{ , } \epsilon \sim \mathcal{G}(0, 1)
\end{align}
where $\mathcal{G}(0, 1)$ is the Gumbel distribution, and $\tau$ is a temperature hyperparameter which increases the  discreteness of samples when $\tau \to 0$.

We will discuss different head selection strategies in Section \ref{subsec:strategy}, which make binary selection decisions based on real-valued posterior $q_{\phi}(z_{t}^{(h)})$.



\subsection{Attention Selection Strategies}
\label{subsec:strategy}

\begin{figure*}[h]
\centering
\begin{minipage}{0.48\textwidth}
\centerline{\includegraphics[width=\linewidth]{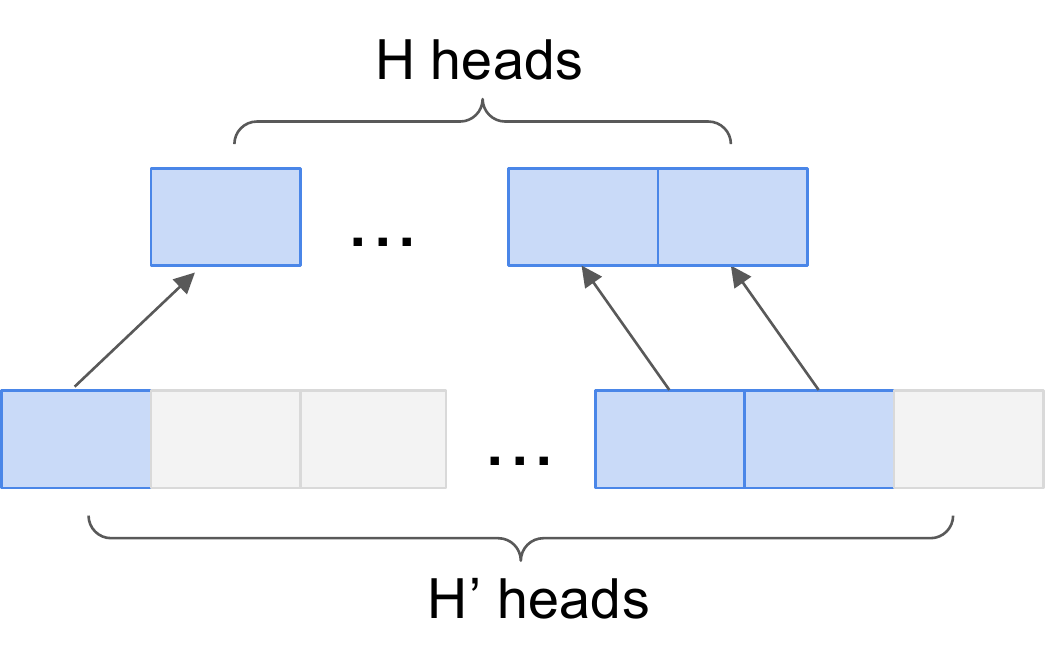}}
\centerline{\small{(a) Subset strategy.}}
\label{fig:subset}
\end{minipage}
\hspace{0.2cm}
\begin{minipage}[c]{0.48\textwidth}
\centerline{\includegraphics[width=\linewidth]{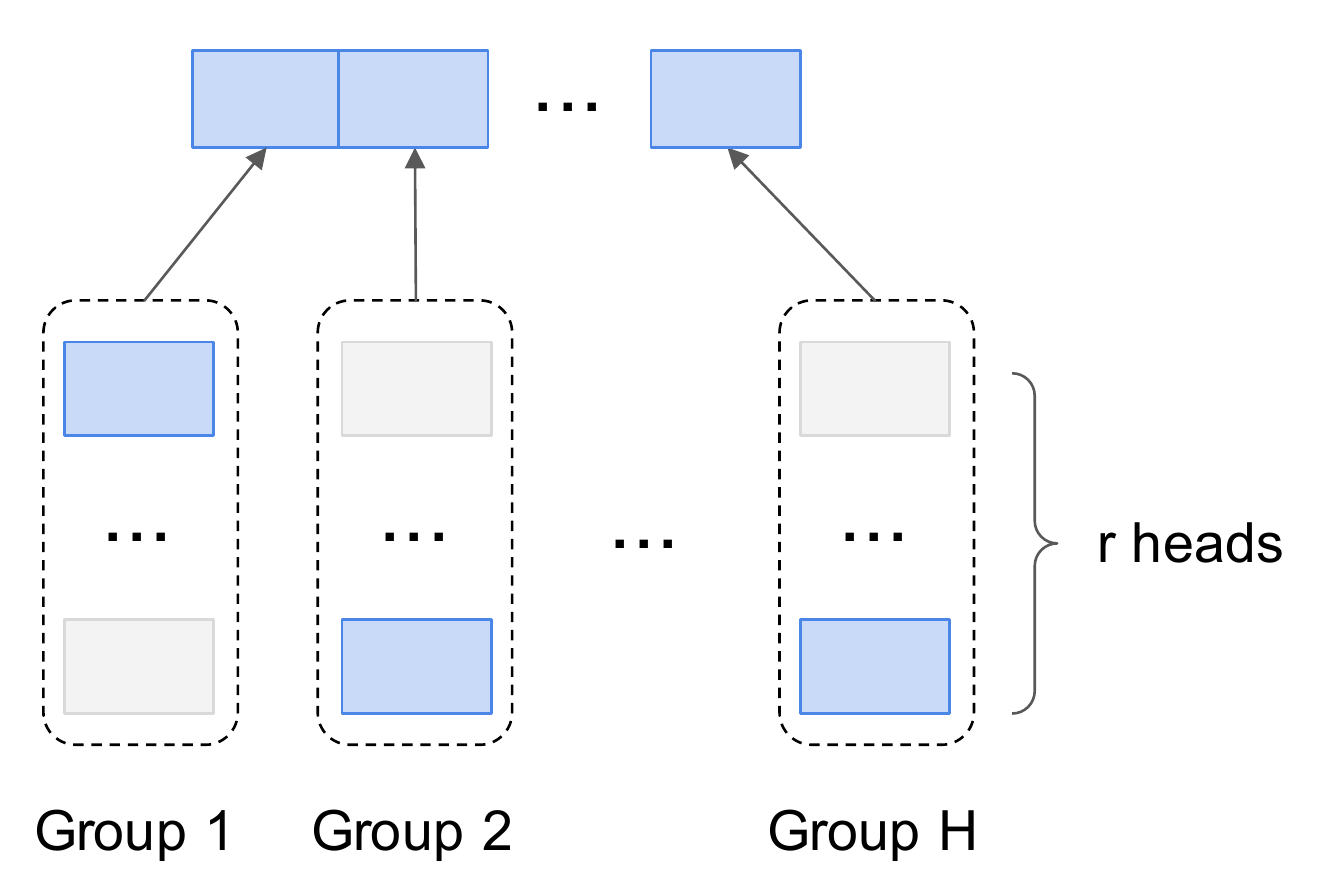}}
\centerline{\small{(b) Group strategy.}}
\label{fig:group}
\end{minipage}
\caption{Attention sharing strategies. The blue heads are selected while the grey heads are not.}
\label{fig:strategy}
\end{figure*}

Suppose that the output dimension of multi-head attention is $d$, and the dimension of each attention head is $\frac{d}{H}$. 
We provide a large pool of $H'$ ($H'>H$) attention head candidates in every Transformer layer, and $H'$ is a hyperparameter controlling the search space size of attention selection strategies.  The model requires attention outputs to have a consistent dimension $d$, so each task needs to select $H$ heads among $H'$ candidates. Knowledge transfer is enabled among tasks accessing the same attention heads, and interference is disabled among tasks without attention sharing. We introduce two strategies for the attention head selection: subset strategy and group strategy.

\textbf{Subset strategy}. 
The strategy is straightforward, and we compare the posterior $\{q_{\phi}(z_{t}^{(h)}): h\in [1,H']\}$ of all $H'$ heads given a task $t$. A subset of $H$ heads with the highest posterior are selected by the task, and there are $C^{H'}_{H}$ subset choices. The subset strategy is described in Fig.~\ref{fig:strategy}(a).
The binary mask $s_{t}^{(h)}$ indicate whether an attention head $h$ is assigned to task $t$.
\begin{equation}
s_{t}^{(h)} = \left\{
\begin{aligned}
1, & \qquad h \in \text{TopH}(\{q_{\phi}(z_{t}^{(h)})\}), \\
0, & \qquad \text{otherwise},
\end{aligned}
\right.
\end{equation}

where $\text{TopH}(\cdot)$ returns the top $H$ heads with the highest values.

The outputs of the selected heads are concatenated as the attention output. Note that the subset strategy does not consider the order of the attention heads. For example, when head $2$ and $3$ are selected, head $2$ contributes to the beginning part of attention output. With head $1$ and $2$ selected, the output of head $2$ goes to the last part of the attention output.



\textbf{Group strategy}. We further propose group strategy to preserve the order of attention heads during head selection.
Different from the subset strategy, the group strategy first divides $H'$ heads into $H$ groups. As is shown in Fig.~\ref{fig:strategy}(b), each group contains $r=\frac{H'}{H}$ candidates. Each task could choose one attention head from each group, and has access to $H$ heads per layer. There are $r^{H}$ possible combinations of heads. The group strategy keeps the head order in that heads from group $g$ only contribute to $g$'s corresponding dimensions in the attention output. The head with the highest posterior in its group would be selected by a given task $t$. We use binary masks $\{s^{(h)}_{t}\}$ to indicate the selection of head $h$ in group $g$. 
\begin{equation}
s_{t}^{(h)} = \left\{
\begin{aligned}
1, & \qquad h = \argmax(\{q_{\phi}(z_{t}^{(h')}): h'\in g\}), \\
0, & \qquad \text{otherwise}.
\end{aligned}
\right.
\end{equation}



The output of group $g$ is:
\begin{align}
    \mathbf{x}^{(g)} = \sum\limits_{h\in g} s^{(h)}_{t} \cdot \mathbf{x}^{(h)}.
\end{align}

The outputs of $H$ groups are concatenated as the output of the attention module for task $t$.

With either subset or group strategy, the sequence model is trained to assign attention heads to different tasks to maximize the lower bound in inequality (\ref{eq:lb}). The number of additional parameters $\{\phi_{t}^{(h)}\}$ introduced by our attention selection is only $T\times H'\times L$, where $T$ is the number of tasks, $H'$ is the number of head candidates per layer, and $L$ is the number of layers. It is small compared with the total parameter size of the model, and head selection is thus lightweight and memory efficient. Moreover, the head selection is inherently a pruning process. Regardless of the size of head candidates, only a fixed number of attention heads are involved in computation for a given task. Hence our approach is also computationally efficient in both training and inference.

\section{Experiments}
\label{sec:experiment}


We evaluate sequence models in multilingual and multi-domain settings respectively. Various applications are considered including multilingual machine translation (MT), automatic speech recognition (ASR) and speech translation (ST) in both multilingual and multi-domain settings. We include widely used sequence models built on multi-head attention as strong baselines. All baselines have an encoder-decoder architecture, leveraging attention in each encoder and decoder layer.

1.Transformer \citep{DBLP:conf/nips/VaswaniSPUJGKP17}. It is a state-of-the-art model in machine translation, which takes texts in source languages as inputs and generates texts in target languages.

2. S2T Transformer \citep{wang2020fairseq}. As a variant of Transformer for speech recognition and translation, S2T Transformer takes audio features and generates target texts. It is a stack of a convolutional subsampler and a Transformer model, where the subsampler processes input features (i.e., log mel-filter features in our experiments) and sends them to Transformer for text generation.

3. Adapter model \citep{bapna2019simple}. Adapters have been shown as an effective approach to language and domain adaptation. Based on a well-trained Transformer or S2T Transformer, task-specific layers are added on top of each encoder and decoder layer. Parameters of adapter layers are trained with other model parameters frozen. A typical adapter layer consists of two feed-forward sub-layers. Adapter is applied to every task in our experiments.

We integrate attention selection strategies into the self-attention\footnote{We also tried head selection in the encoder-decoder attention but did not observe big improvements when using it alone or in combination with self-attention head selection.} module.
Our implementation is based on the \textsc{fairseq} toolkit \citep{ott2019fairseq,wang2020fairseqs2t}.


\subsection{Machine Translation}

The task of machine translation is to translate a text from the source language into the target language. BLEU is an evaluation metric in translation, which measures the overlap between model translations and the ground truth \citep{papineni2002bleu}. Higher BLEU reflects better translation quality.

\textbf{Dataset}. We experiment with public multilingual machine translation datasets collected by WMT shared tasks as used by \citep{liu2020multilingual}. The dataset consists of parallel sentences between English and other $14$ languages\footnote{The 14 languages are: Chinese, Czech, Estonian, Finnish, French, German, Gujarati, Kazakh, Latvian, Lithuanian, Romanian, Russian, Spanish, Turkish}. Its data statistics are summarized in Appendix \ref{app:data}. We evaluate models on both one-to-many (O2M) and many-to-one (M2O) translations, which are translation from English to $14$ languages and from $14$ languages to English respectively.


\textbf{Model configurations}. All models have $6$ encoder layers and $6$ decoder layers with $4$ attention heads per layer (i.e., $H=4$). The embedding dimension is $512$ and the feed-forward dimension is $1024$. They are trained with a batch size of $131$k tokens and a learning rate of $0.0007$. For O2M translation, attention selection models and Transformer are trained for $140$k steps. As for M2O translation, they are trained for $100$k steps. The attention selection is based on the source language in the encoder side for M2O translation, and is based on the target language in the decoder part for O2M translation.

The adapter model is initialized with parameters from the trained Transformer. It then trains the new parameters of the adapter layers for $40$k steps with Transformer parameters frozen.
The adapter layers are added to Transformer for each language direction, and they have an intermediate dimension of $256$. The dimension is selected so that the number of parameters (460M) in the adapter model is close to the parameter size (420M) in attention selection models.
Our attention selection sets the number of attention head candidates as $8$ in each layer (i.e., $H'=8$) for both subset and group strategies. We will discuss how the hyperparamter $H'$ affects model performance in Section~\ref{sec:discussion}. 


\begin{table*}[htbp!]
\caption{BLEU ($\uparrow$) of Machine Translation on WMT Datasets (AVG-A: average BLEU over $14$ directions, High and Low are average BLEU over high- and low-resource languages respectively.)}
\label{tab:results_mt}
\centering
\begin{tabular}{ccccccccc}
\thickhline
 & & \multicolumn{3}{c}{O2M} & & \multicolumn{3}{c}{M2O} \\ \cline{3-5} \cline{7-9}
 & & AVG-A & High & Low & & AVG-A & High & Low \\ \hline
Transformer & & 20.1 & 25.7 & 16.0 & & 22.8 & 27.9 & 19.0 \\
Adapter & & 20.9 & 26.7 & \textbf{16.6} & & 23.3 & 28.7 & 19.3 \\ \hline\hline
Group strategy & & \textbf{21.0} & \textbf{27.1} & 16.4 & & \textbf{23.5} & \textbf{28.8} & \textbf{19.6} \\
Subset strategy & & 20.9 & 27.0 & 16.4 & & 23.3 & 28.7 & 19.4 \\ 
 \thickhline
\end{tabular}
\end{table*}

\textbf{Results}. We group $14$ language directions based on their amount of training data. We have $6$ high-resource languages with more than $10$M parallel sentences, 
$8$ low-resource languages with fewer than $10$M sentence pairs.
Table~\ref{tab:results_mt} shows model performance on WMT datasets. Both attention selection and adapter models demonstrate gains over the multilingual Transformer. Group strategy achieves  +$0.9$ and +$0.7$ BLEU on average of $14$ language directions  in O2M and M2O translations respectively. Adapter has comparable performance to both group and subset strategies. We note that adapter increases computational costs due to the additional $12$ adapter layers added to Transformer, while the attention head selection approaches preserve the computation efficiency.



\subsection{Speech Recognition}
\label{subsec:asr}

The task of Automatic Speech Recognition (ASR) is to transcribe source audios in the same language. Word error rate (WER) is ASR evaluation metric, which measures the difference of model outputs from the ground truth \citep{klakow2002testing}. Lower WER indicates better recognition.

\noindent\textbf{Model configuration}. Models included in the experiments of speech recognition are S2T Transformer, S2T Transformer with adapter layers and S2T Transformer with attention selection. Following the setup of \citep{salesky2021mtedx}, all models have $1024$ channels in the input convolutional subsampler, $12$ encoder layers and $6$ decoder layers with $4$ attention heads per layer. The embedding dimension is $256$ and the feed-forward dimension is $2048$. We set a batch size of $320$k tokens and a learning rate of $0.0005$ during training. 
Attention selection models and S2T Transformer are trained for $250$ epochs. Adapter model is initialized with parameters of the trained S2T Transformer, and is then trained for another $200$ epochs with only adapter layer parameters tuned. The intermediate dimension of adapter layers is again set as $256$.
To prevent over-fitting, we stop the model training when the model does not improve on the  validation set for $10$ epochs. To reduce the performance variance, we average checkpoints of the last $10$ epochs, and use the averaged model for evaluation.

\subsubsection{Multilingual Speech Recognition}

\textbf{Dataset}. We use the multilingual TEDx (mTEDx) dataset for speech recognition \citep{salesky2021mtedx}. It collects audio recordings from TEDx talks. Eight languages are covered including Arabic (ar), German (de), Greek (el), Spanish (es), French (fr), Italian (it), Portuguese (pt) and Russian (ru).



\begin{table*}[htbp!]
\centering
\caption{WER ($\downarrow$) of Speech Recognition on mTEDx Dataset}
\label{tab:asr}
\begin{tabular}{cccccccccc}
\thickhline
 & AVG & ar & de & el & es & fr & it & pt & ru \\ \hline
S2T Transformer & 49.0 & 109.5 & 72.3 & 43.3 & 23.9 & 27.8 & 28.6 & 31.0 & 55.3 \\
Adapter & 41.1 & 93.4 & 57.2 & 33.0	& 21.4 & 25.3 & 24.3 & 27.2 & 46.7 \\ \hline\hline
Group strategy & \textbf{40.0} & 94.2 & 59.8 & 33.5 & 18.2 & 22.0 & 21.9 & 24.6 & 45.5 \\
Subset strategy & 44.7 & 97.3 & 65.3 & 38.7 & 22.4 & 25.8 & 26.4 & 29.0 & 52.4 \\ \thickhline
\end{tabular}
\end{table*}


\textbf{Results}. S2T transformer share all parameters among languages. Attention selection models select attention heads based on the source and target languages in the multilingual setting. Adapter adds adapter layers based on the language directions.
We report the ASR results in Table~\ref{tab:asr}. Attention selection with either group or subset strategy is shown to reduce the word error rate in comparison with S2T Transformer. Adapter model also achieves lower WER than S2T Transformer. Group strategy yields the lowest WER, achieving an average drop of $18.4\%$ compared with S2T Transformer. Adapter reduces the WER of S2T Transformer by $16.1\%$.

\subsubsection{Multi-Domain Speech Recognition}

\textbf{Dataset}. Besides mTEDx data, we include two other public datasets, CoVoST and EuroParl, which are commonly used for speech translation. Since source audios are accompanied by transcripts, we could use their source audio-text data for speech recognition tasks. We investigate multi-domain modeling with these three datasets.

1. CoVoST \citep{Wang2020CoVoST2A}. With Common Voice as the audio source, CoVoST covers speech-to-text translations from $22$ languages to English and from English to $15$ languages. \\
2. EuroParl \citep{iranzo2020europarl}. It provides paired audio-text instances from and into $6$ European languages, which are compiled from the debates in European Parliament. 

\begin{table*}[htbp!]
\caption{WER ($\downarrow$) of Speech Recognition on mTEDx, CoVoST and EuroParl Dataset}
\label{tab:asr_domain}
\centering
\begin{tabular}{cccc}
\thickhline
 & mTEDx & CoVoST & EuroParl \\ \hline
S2T Transformer (separate) & 49.0 & 41.9 & 115.0 \\ 
S2T Transformer (joint) & 42.7 & 38.3 & 25.6 \\
Adapter & 41.7 & 37.0 & \textbf{24.0} \\ \hline\hline
Group strategy & \textbf{41.0} & \textbf{36.4} & 24.3 \\
Subset strategy & 41.8 & 37.0 & 25.0  \\
\thickhline
\end{tabular}
\end{table*}


\textbf{Results}. In the multi-domain setting, attention selection models assign different heads to each domain, and adapter model adds domain-specific adapter layers to S2T Transformer. We train models for $400$ epochs in the multi-domain setting. Table~\ref{tab:asr_domain} reports WER of models in three domains: mTEDx, CoVoST and EuroParl respectively. The S2T Transformer jointly trained on multi-domain data (in the row of ``S2T Transformer (joint)'') reduces WER by $12.9\%$, $8.6\%$ and $77.7\%$ in three domains respectively, when compared with the models separately trained in individual domains (in the row of ``S2T Transformer (separate)''). This demonstrates the benefits of positive transfer between domains. 

The performance of speech recognition could be further improved by the mitigation of the domain interference. Both attention selection and adapter model achieve lower WER than the joint S2T Transformer. Attention selection with group strategy has the lowest WER on both mTEDx and CoVoST datasets, decreasing WER by $4.0\%$ and $5.0\%$
respectively in comparison with joint S2T Transformer. The best system on EuroParl is adapter model, yielding a WER reduction by $6.3\%$ than the joint S2T Transformer.

\subsection{Speech Translation}

Now with a focus on the task of speech translation, we again design experiments in multilingual and multi-domain settings. In the multilingual setup, we train translation models with samples in multiple languages to investigate language interference. As for the multi-domain setup, the models are trained with data from multiple domains so that we could look into the domain interference. BLEU serves as the evaluation metric of speech translation systems. 

\textbf{Baselines}. We use the same baselines as in speech recognition. As recommended by \citep{salesky2021mtedx}, we initialize the encoders in speech translation with the encoders trained in the task of speech recognition in Section ~\ref{subsec:asr} for the purpose of improving training efficiency and performance.

\textbf{Model configurations}. All models are trained for up to $400$ epochs. Other model configurations in ST are the same as those in ASR.

\subsubsection{Multilingual Speech Translation}

To explore language interference, we perform experiments on multilingual speech translation.

\textbf{Dataset}. We again use mTEDx dataset for multilingual speech translation. Besides speech recognition data, mTEDx also collects speech translation data from TEDx talks. Its test set covers $13$ language directions. The training data is provided in $10$ of these directions, so there are $3$ zero-shot directions.

\begin{wraptable}{l}{0.6\linewidth}
  \centering
\caption{BLEU ($\uparrow$) of Speech Translation on mTEDx (AVG-A: average over all directions, AVG-T: average of $10$ training directions, and AVG-Z: average of $3$ zero-shot directions)}
\label{tab:st_lang}
\centering
\begin{tabular}{cccc}
\thickhline
 & AVG-A & AVG-T & AVG-Z \\ \hline
S2T Transformer & 13.2 & 14.6 & 8.5 \\ 
Adapter & - & 14.8 & - \\ \hline\hline
Group strategy & \textbf{15.2} & \textbf{16.7} & \textbf{10.4} \\
Subset strategy & 13.3 & 14.7 & 8.5 \\ 
\thickhline
\end{tabular}
\end{wraptable}

\textbf{Results}. Table~\ref{tab:st_lang} summarizes the multilingual speech translation results on mTEDx. Since adapter model brings in language-specific layers, it cannot deal with zero-shot translations. Both attention selection models and adapter model bring improvements over S2T Transformer which is jointly trained in $13$ language directions. It suggests that multiple languages interfere within S2T Transformer whose parameters are shared by all languages. Attention selection with group strategy achieves the best translation performance. In comparison with S2T Transformer, group strategy achieves an average of +$2.1$ and +$1.9$ BLEU in training and zero-shot directions respectively. It leads to +$2.0$ BLEU on average of all directions.


\subsubsection{Multi-Domain Speech Translation}

In this experiment, we investigate interference across domains in the task of speech translation, and evaluate the effectiveness of different models in multi-domain training. The attention selection now is based on the data domain instead of languages, i.e., samples in different domains would choose their own attention heads. Similarly for adapter model, its adapter layers are domain-specific in this setup.

We again use CoVoST and EuroParl as additional domains. We focus on the $13$ language directions in mTEDx test set, and use the subset of CoVoST and EuroParl corpora in the same directions. CoVoST has $5$ common directions\footnote{\{es, fr, it, pt, ru\}-en} and EuroParl has $11$ common directions\footnote{es-\{en, fr, it, pt\}, fr-\{en, es, pt\}, it-\{en, es\}, pt-\{en, es\}} as mTEDx. Details about these datasets are included in Appendix \ref{app:data}. 

\begin{table*}[htbp!]
\caption{BLEU ($\uparrow$) of Speech Translation on mTEDx, CoVoST and EuroParl Dataset}
\label{tab:st_domain}
\centering
\begin{tabular}{ccccccccc}
\thickhline
 & & \multicolumn{3}{c}{TEDx} & & CoVoST & & EuroParl \\ \cline{3-5}\cline{7-7}\cline{9-9}
 & & AVG-A & AVG-T & AVG-Z & & AVG & & AVG \\ \hline
S2T Transformer (separate) & & 13.2 & 14.6 & 8.5 & & 17.6 & & 19.1 \\ 
S2T Transformer (joint) & & 13.7 & 13.6 & 13.9 & & 17.3 &  & 19.0 \\
Adapter & & 14.0 & 14.3 & 13.2 & & 17.9 & & 20.0 \\ \hline\hline
Group strategy & & \textbf{15.6} & \textbf{15.9} & \textbf{14.8} & & \textbf{19.6} & & \textbf{20.8} \\
Subset strategy & & 13.8 & 14.3 & 13.1 & & 17.9 & & 19.2  \\ 
\thickhline
\end{tabular}
\end{table*}

\textbf{Results}. Table~\ref{tab:st_domain} shows the average BLEU of speech translation in mTEDx, CoVoST and EuroParl. We report results in the row of ``S2T Transformer (joint)'' when S2T Transformer is trained with a mixture of three datasets. The results are included in the row ``S2T Transformer (separate)'' when S2T Transformer is trained on each dataset independently. Zero-shot translations in mTEDx benefit a lot from additional data of CoVoST and EuroParl, as the joint S2T Transformer shows an average of +$5.4$ BLEU over separate S2T Transformer. However, there is a drop of $1.0$ BLEU in its training directions, brought by the interference from CoVoST and EuroParl domains. 

Again we observe that both attention selection and adapter model bring gains to the joint model in individual domains. Compared with the joint S2T Transformer, adapter model improves mTEDx translation by $0.3$ BLEU, CoVoST translation by $0.6$ BLEU and EuroParl by $1.0$ BLEU on average. The attention selection with group strategy outperforms all other models. Its average BLEU gain over adapter model is $1.6$ BLEU in mTEDx, $1.7$ BLEU in CoVoST and $0.8$ in EuroParl.



\section{Discussion}
\label{sec:discussion}


\textbf{Hyperparameter $H'$}. The attention selection models set a hyperparameter $H'$ as the total number of attention head candidates in multi-head attention, which controls the search space of attention sharing strategies. We now explore how the performance varies with $H'$ for group and subset strategies.

\begin{wrapfigure}{r}{0.45\textwidth}
  \begin{center}
    \includegraphics[width=0.45\textwidth]{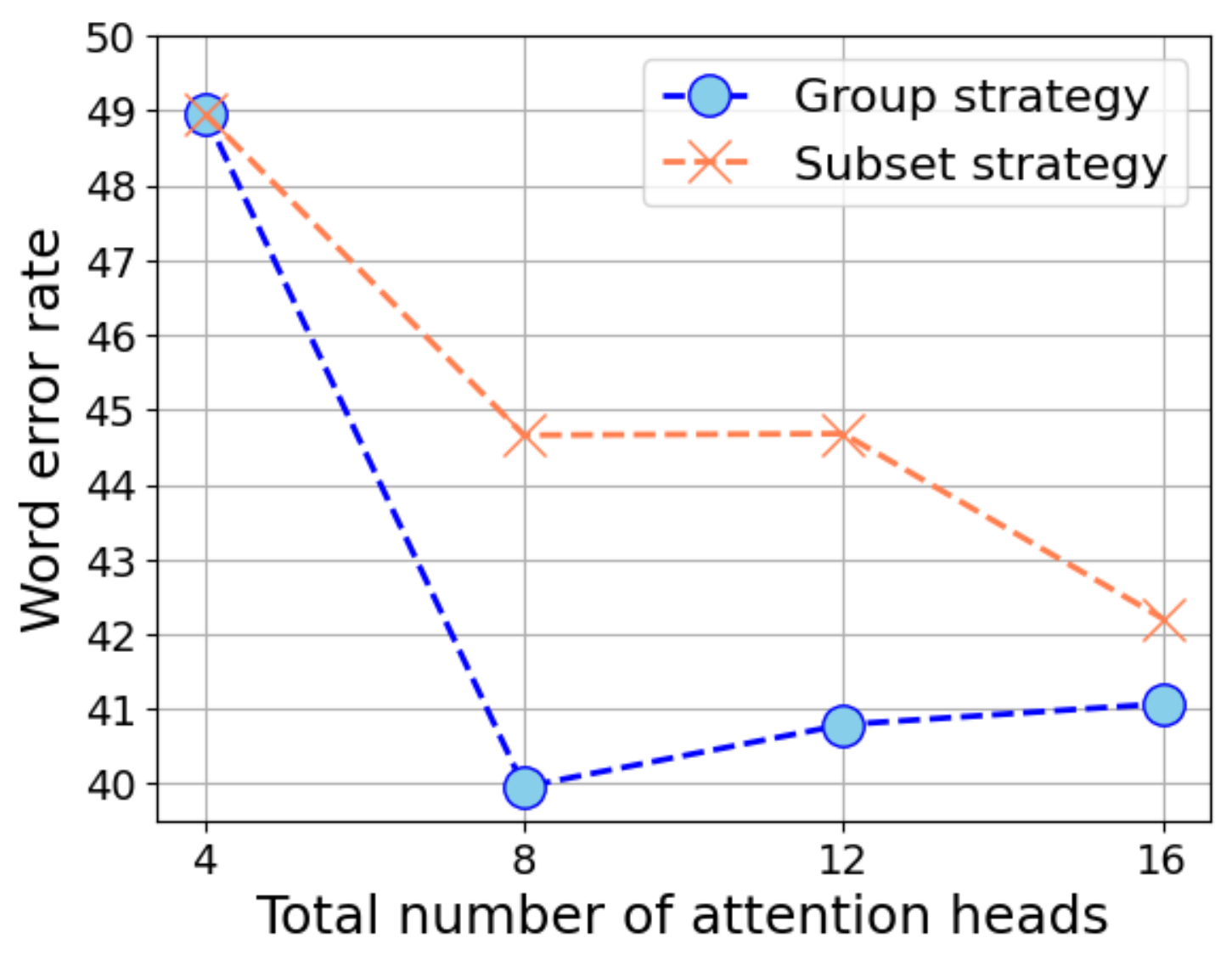}
  \end{center}
  \caption{WER of speech recognition on mTEDx with different number of candidates.}
  \label{fig:discuss_head_num}
\end{wrapfigure}


Evaluated on the task of multilingual speech recognition, models have the same hyperparameters as those in multilingual ASR experiments except for $H'$. Attention selection models are configured with $H'=4, 8, 12, 16$ respectively, and Figure~\ref{fig:discuss_head_num} shows the change of WER with $H'$. 

When $H'=4$, there is no attention selection and all attention heads are shared by different languages. We observe a large drop of error rate as $H'$ increases from $4$ to $8$.
For the subset strategy, WER keeps decreasing when the number of head candidates grows from $4$ to $16$.
As for group strategy, $H'=8$ is the optimal hyperparameter on the ASR task. As we continue increasing $H'$ to $12$ and $16$, the error rate increases a bit. The performances of subset and group strategies are close when $H'=16$.

The search space of group strategy is a strict subset of the space of subset strategy. However, we observe that group strategy shows comparable or better performance than subset strategy across tasks, including MT, ASR and ST. One possible explanation is that group strategy keeps the head order information while subset strategy does not. With a larger pool of head candidates, there is less sharing among tasks. The performance of the group strategy degrades a bit due to less positive transfer dependent on attention sharing. As for the subset strategy, better head assignments are learned with enlarged search space. 

\begin{figure}[htbp!]
\begin{center}
\begin{minipage}[b]{0.49\linewidth}
        \centering
        \includegraphics[width=1.0\columnwidth]{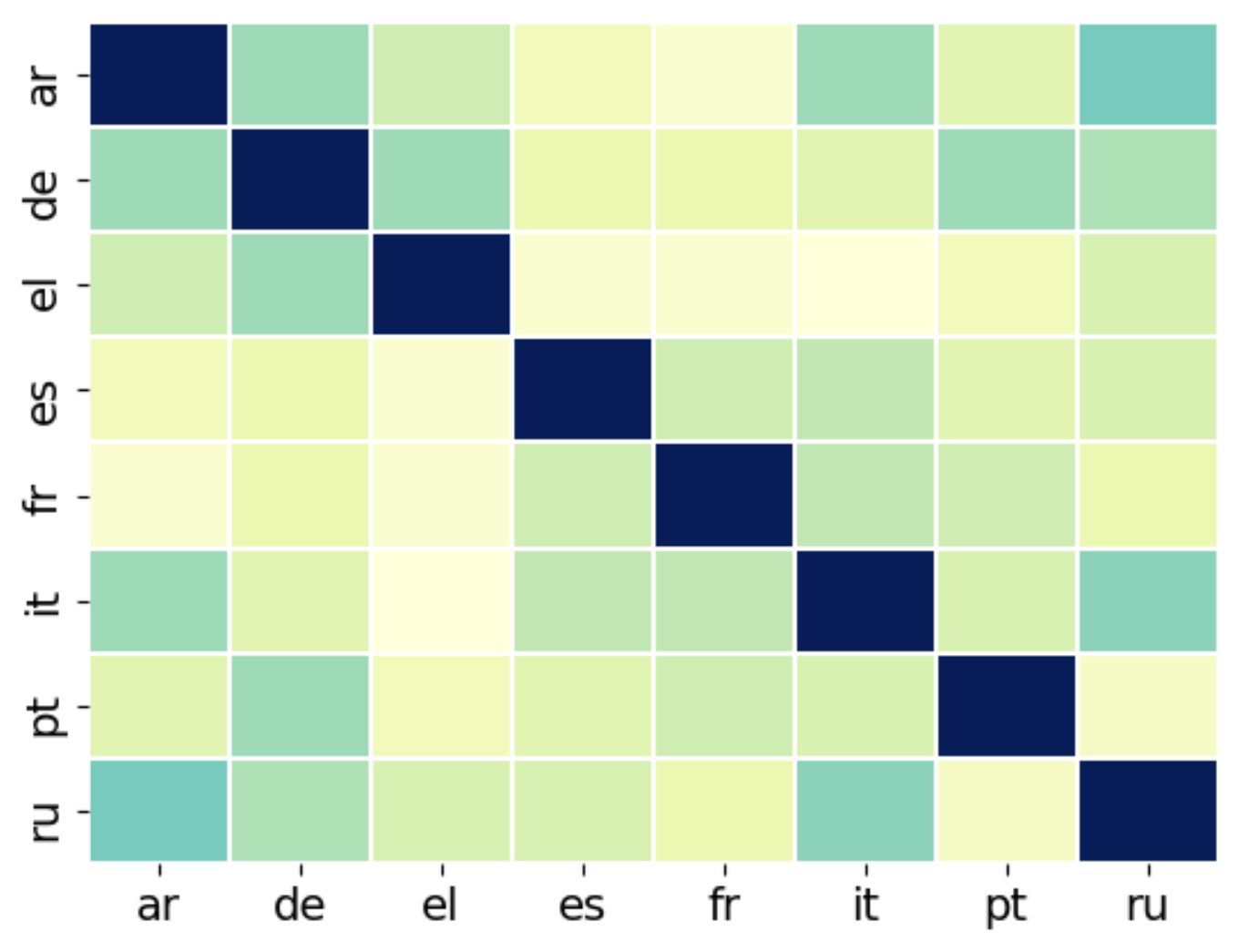}
        {(a) Encoder with group strategy.}
\end{minipage}
\begin{minipage}[b]{0.49\linewidth}
        \centering
        \includegraphics[width=1.0\columnwidth]{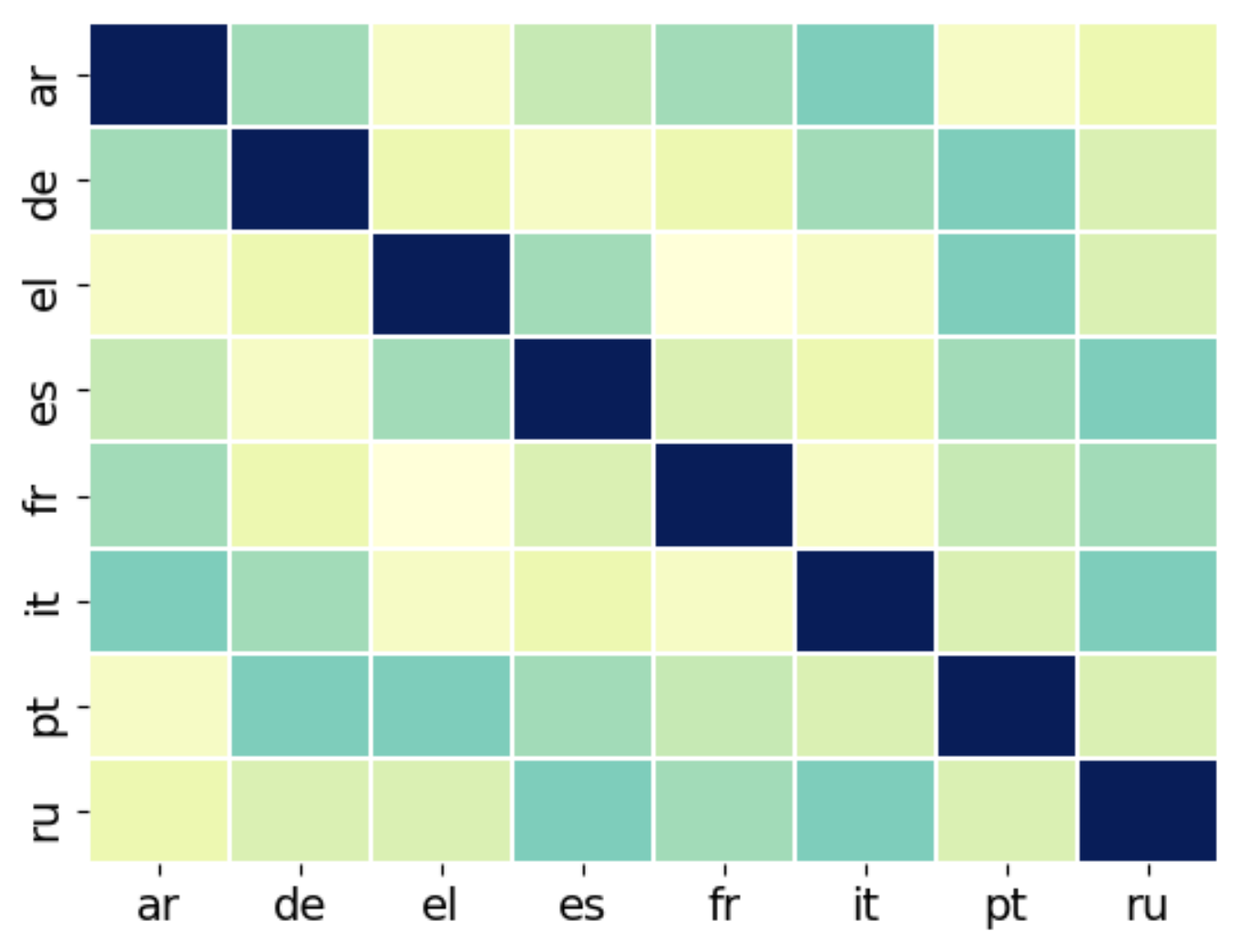}
        {(b) Decoder with group strategy.}
\end{minipage}
\caption{Heatmap to visualize the sharing between languages in multilingual ASR (The darker a language pair is, the more attention heads they share.)}
\label{fig:lang_heatmap}
\end{center}
\end{figure}

\textbf{Attention Sharing among Languages}. We now analyze the attention sharing pattern among languages. Take the multilingual model on mTEDx speech recognition as an example, whose head selection is learned with group strategy. We count the number of heads shared by each language pair in the model, and visualize it with a heatmap in Fig.~\ref{fig:lang_heatmap}, where the darkness reflects the amount of sharing. The diagonal cells in the heatmap corresponds to the number of attention heads used by each language, i.e., the total number of attention heads in all layers.

For European languages including Spanish (es), French (fr), Italian (it) and Portuguese (pt), their shared attention heads are fewer in decoder than in encoder. This seems contradicted with previous findings that parameter sharing is beneficial for languages with high linguistic proximity. We note that they are high-resource languages in mTEDx corpus, which is also justified by their relatively lower WER. Their data is sufficient to learn good speech recognition, and sharing parameters with other languages hurt the preservation of the language specificity. This explains why the high-resource European languages do not share too many heads in the learned group strategy.

Another pattern we observe from Fig.~\ref{fig:lang_heatmap} is that low-resource languages tend to share more attention heads with high-resource languages. For example, Arabic (ar) and Russian (ru) have relatively more sharing with Italian (it) than other languages. Low-resource languages benefit from the knowledge transfer from high-resource languages.

\begin{figure}[htbp!]
\begin{center}
\begin{minipage}[b]{0.49\linewidth}
        \centering
        \includegraphics[width=1.0\columnwidth]{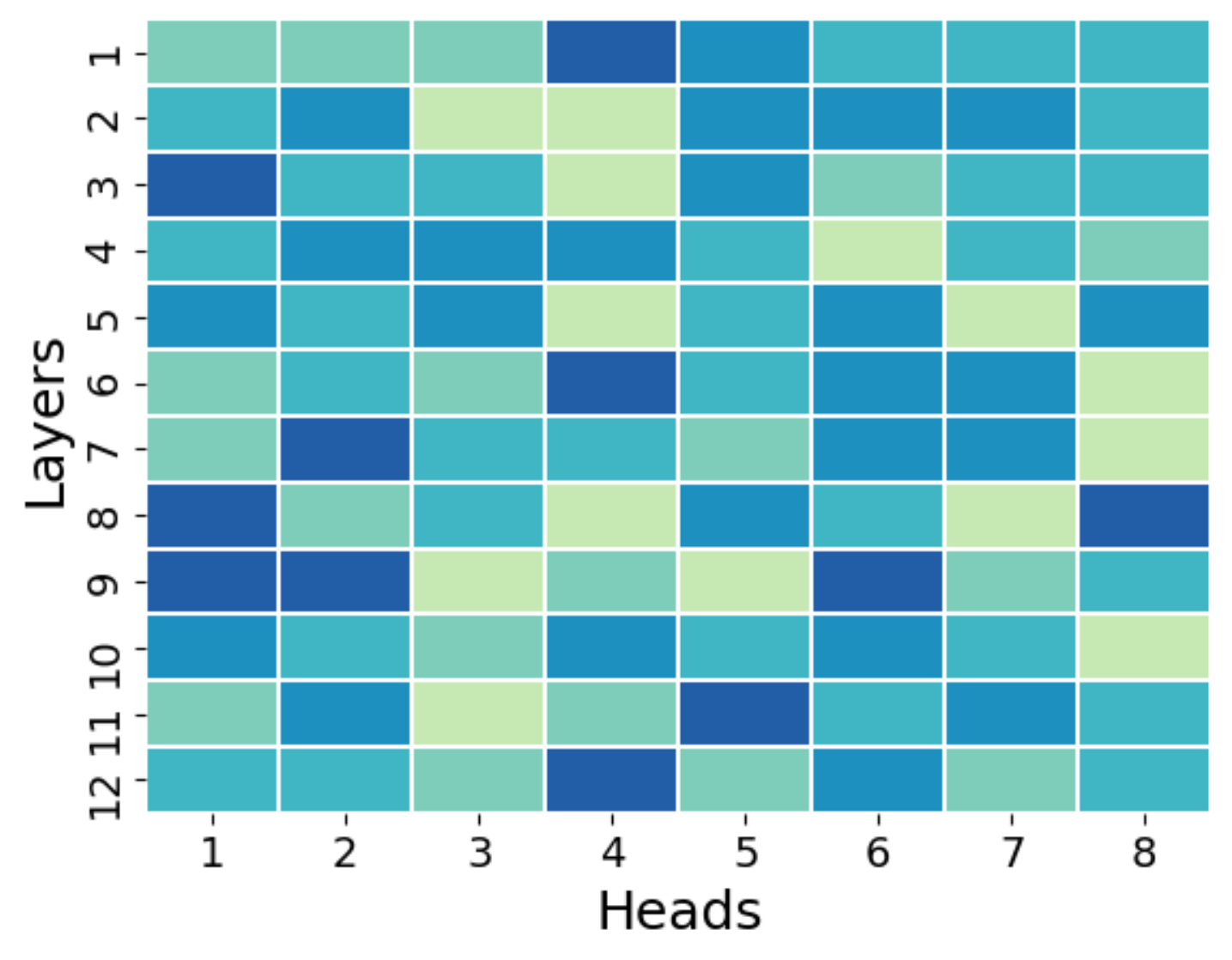}
        {(a) Encoder attention with subset strategy.}
\end{minipage}
\begin{minipage}[b]{0.49\linewidth}
        \centering
        \includegraphics[width=1.0\columnwidth]{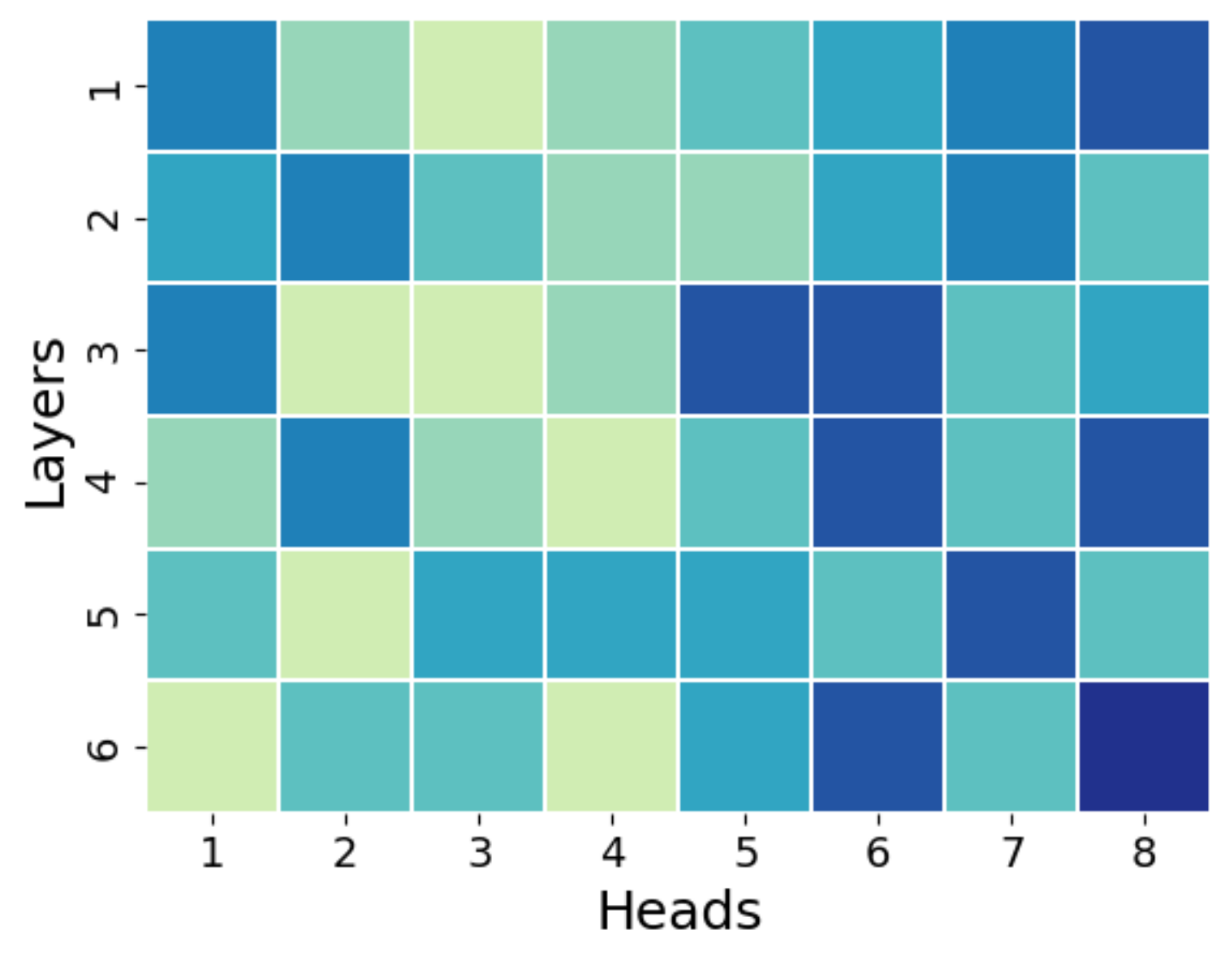}
        {(b) Decoder attention with subset strategy.}
\end{minipage}
\begin{minipage}[b]{0.49\linewidth}
        \centering
        \includegraphics[width=1.0\columnwidth]{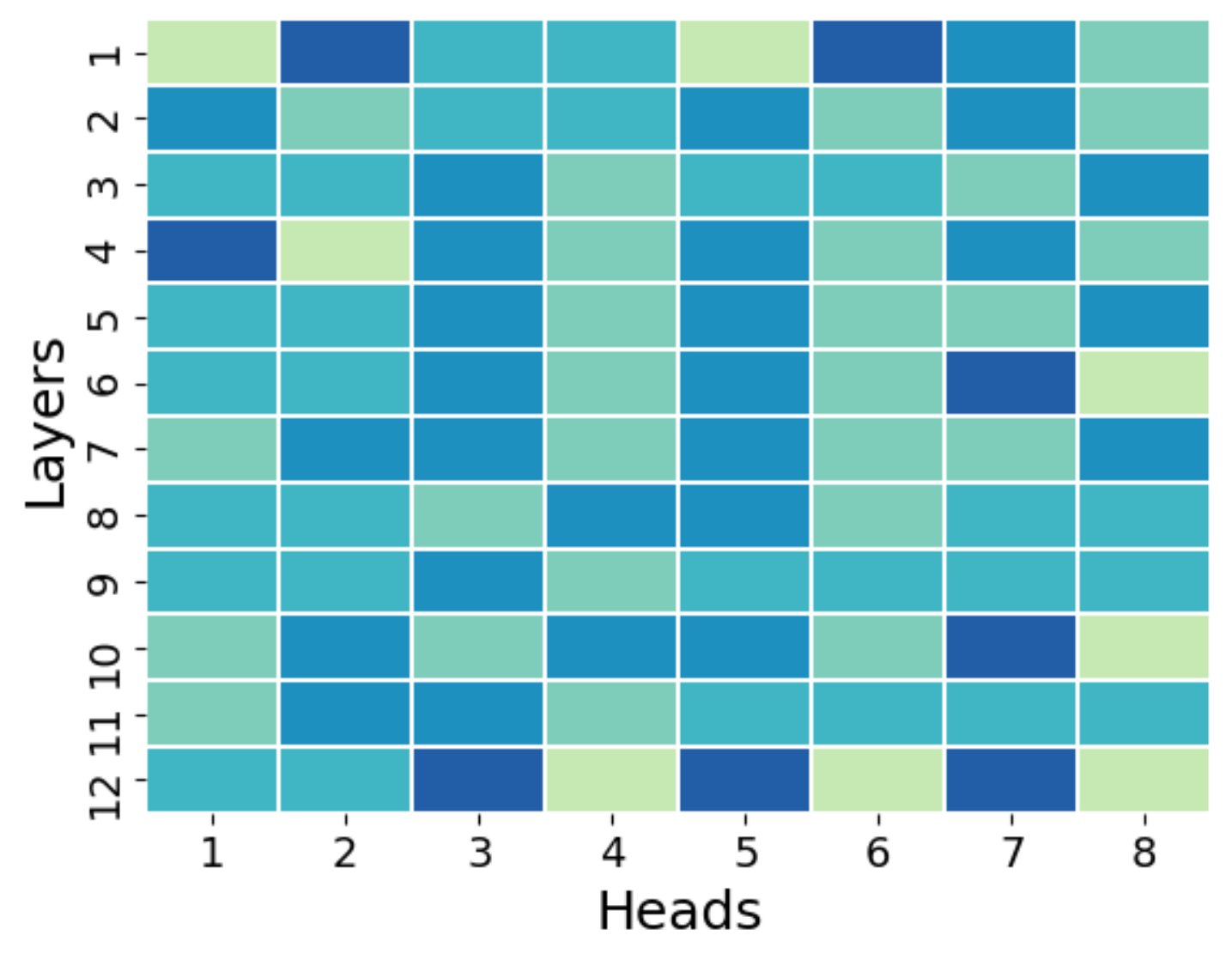}
        {(c) Encoder attention with group strategy.}
\end{minipage}
\begin{minipage}[b]{0.49\linewidth}
        \centering
        \includegraphics[width=1.0\columnwidth]{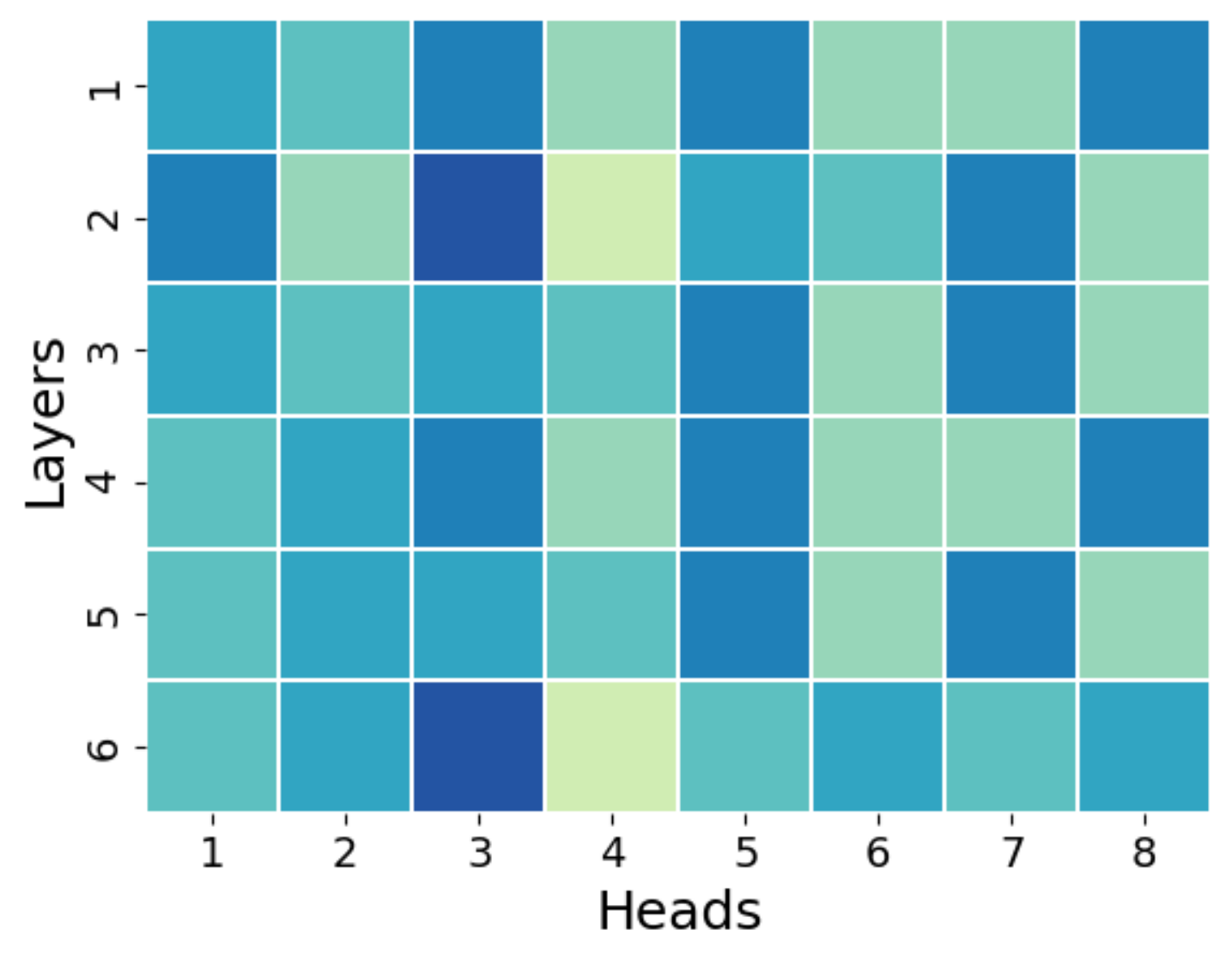}
        {(d) Decoder attention with group strategy.}
\end{minipage}
\caption{Heatmap to visualize the load of attention heads (The darker a head is, the more languages it supports.}
\label{fig:attn_heatmap}
\end{center}
\end{figure}

\textbf{Load of Attention Heads}. To gain an insight into the load of attention heads, we analyze how many languages an attention head is shared by. With both subset and group strategies, we look into attention heads in encoder and decoder respectively. We study ASR models which learn language based attention selection on mTEDx data containing $8$ languages.
The language load of each attention head is measured by the number of languages sharing the given head. Fig.~\ref{fig:attn_heatmap} visualizes the load of attention heads in each layer with a heatmap. The darkness reflects the load of an attention head.

By comparing encoder heads in Fig.~\ref{fig:attn_heatmap}(a) and (c), we note that group strategy results in more balanced load among attention heads than subset strategy, as there is less color variation in the heatmap of group strategy. Similar pattern could be observed in decoder, and decoder attention heads have more balanced load with group strategy.

Now we compare the load of attention heads across layers. With subset strategy, the load imbalance is observed in heads of almost every encoder and decoder layer from the color contrast in the heatmap. As for group strategy, the load is more balanced in heads of middle layers (i.e., encoder layers $5-9$ and decoder layers $3-5$) than those in bottom and top layers.

\textbf{Head Selection in Encoder and Decoder}. In our experiments, attention selection is applied to both encoder and decoder in ASR and ST experiments, considering that both encoder and decoder handle multiple languages.
We want to measure how the model performance is affected by attention selection in encoder and decoder respectively. Take the multilingual ASR as an example, Table ~\ref{tab:enc_dec_asr} reports the word error rate of models which enable attention selection in encoder only, in decoder only as well as in both encoder and decoder. We set the same hyperparameters as used in the experiment of multilingual ASR. When the attention selection is applied to encoder (or decoder) only, $4$ attention heads are shared by all languages in each decoder (or encoder) layer.

\begin{table}[htbp!]
\centering
\caption{Ablation Study in  WER of Multilingual Speech Recognition on mTEDx}
\label{tab:enc_dec_asr}
\begin{tabular}{ccccc}
\thickhline
Component with attention selection & & Encoder only & Decoder only & Encoder+Decoder \\ \hline
Group strategy & & 42.2 & 46.2 & 40.0 \\
Subset strategy & & 45.4 & 47.5 & 44.7 \\ \thickhline
\end{tabular}
\end{table}

As is shown in Table~\ref{tab:enc_dec_asr}, attention selection in only encoder (c.f. column `Encoder only`'') or decoder (c.f. column ``Decoder only'') would increase the word error rate in comparison with the model with attention selection in both encoder and decoder (c.f. column ``Encoder+Decoder''). We also note that attention selection in encoder achieves lower WER than that in decoder.




\section{Conclusion}
\label{sec:conclusion}


Research efforts in multilingual and multi-domain modeling have been driven by the increasing need to improve data efficiency and model performance. In this work, we propose head selection strategies to allow heads to be shared or specialized for different languages or domains. It effectively mitigates interference within multi-head attention which is a core part of strong sequence models, and demonstrates good empirical gains in various text generation tasks.

This work has several limitations left for future research. We did not explore head selection based on both language and data domain. We did not analyze model fairness and robustness. As a technology used for text generation, the model might have systemic bias or produce inappropriate outputs. 


\bibliographystyle{plainnat.bst}
\bibliography{custom}

\newpage

\appendix

\section{Appendix}

\subsection{Dataset}
\label{app:data}

Table~\ref{tab:app_wmt_data} summarizes the number of parallel sentences for $14$ languages in WMT shared tasks.
Table~\ref{tab:app_asr_data} covers three datasets, CoVoST, EuroParl and mTEDx on ASR task, and reports their number of utterances in $8$ languages.
Table~\ref{tab:app_st_data} shows data sizes of three ST datasets including CoVoST, EuroParl and mTEDx. It reports the number of utterances in $13$ language directions.

\begin{table*}[htbp!]
\centering
\caption{Data Statistics of WMT Datasets}
\label{tab:app_wmt_data}
\begin{tabular}{ccc|cccc}
\hline
Language & Code & Size & Language & Code & Size \\ \hline
Gujarati & gu & 10k & Kazakh & kk & 91k  \\
Turkish & tr & 207k  & Romanian & ro & 608k  \\
Estonian & et & 1.94M & Lithuanian & lt & 2.11M   \\
Finnish & fi & 2.66M & Latvian & lv & 4.50M   \\
Czech & cs & 11M & Spanish & es & 15M   \\
Chinese & zh & 25M & German & de & 28M  \\
Russian & ru & 29M & French & fr & 41M  \\ \hline
\end{tabular}
\end{table*}

\begin{table*}[htbp!]
\centering
\caption{Data Statistics of Speech Recognition Task (\# of Utterances)}
\label{tab:app_asr_data}
\begin{tabular}{c|ccc|ccc|ccc}
\hline
Data & \multicolumn{3}{c|}{CoVoST} & \multicolumn{3}{c|}{EuroParl} & \multicolumn{3}{c}{mTEDx} \\ \hline
Split & Train & Dev & Test & Train & Dev & Test & Train & Dev & Test \\ \hline
ar & 2,283 & 1,758 & 1,695 & - & - & - & 11,442 & 1,079 & 1,066 \\
de & 127,577 & 13,503 & 13,503 & 13,099 & 2,653 & 2,644 & 6,659 & 1,172 & 1,126 \\
el & - & - & - & - & - & - & 12,521 & 982 & 1,027\\
es & 78,958 & 13,203 & 13,204 & 7,537 & 1,951 & 1,831 & 99,660 & 905 & 1,012\\
fr & 207,286 & 14,755 & 14,750 & 13,006 & 1,593 & 1,848 & 114,488 & 1,036 & 1,059 \\
it & 31,638 & 8,877 & 8,892 & 11,649 & 1,414 & 1,763 & 48,089 & 931 & 999 \\
pt & 9,158 & 3,315 & 4,021 & 4,977 & 1,794 & 2,292 & 88,123 & 1,013 & 1,020 \\
ru & 12,112 & 6,110 & 6,300 & - & - & - & 28,627 & 973 & 1,132\\ \hline
\end{tabular}
\end{table*}

\begin{table*}[htbp!]
\centering
\caption{Data Statistics of Speech Translation Task (\# of Utterances)}
\label{tab:app_st_data}
\begin{tabular}{c|ccc|ccc|ccc}
\hline
Data & \multicolumn{3}{c|}{CoVoST} & \multicolumn{3}{c|}{EuroParl} & \multicolumn{3}{c}{mTEDx} \\ \hline
Split & Train & Dev & Test & Train & Dev & Test & Train & Dev & Test \\ \hline
el-en & - & - & - & - & - & - & 4,215 & 938 & 1,024 \\
es-en & 78,958 & 13,203 & 13,204 & 7,403 & 1,947 & 1,816 & 35,186 & 899 & 1,001 \\
es-fr & - & - & - & 4,673 & 1,115 & 1,082 & 3,549 & 904 & 1,005 \\
es-it & - & - & - & 4,476 & 1,065 & 1,079 & 5,530 & 16 & 262 \\ 
es-pt & - & - & - & 4,727 & 1,141 & 1,089 & 20,467 & 898 & 1,002 \\
fr-en & 207,286 & 15,560 & 14,952 & 12,446 & 1,481 & 1,804 & 29,634 & 1,035 & 1,058 \\
fr-es & - & - & - & 7,857 & 1,072 & 1,098 & 20,407 & 1,034 & 1,057 \\
fr-pt & - & - & - & 8,183 & 1,048 & 1,100 & 13,047 & 1,035 & 1,058 \\
it-en & 31,638 & 9,095 & 8,937 & 11,285 & 1,400 & 1,686 & - & 929 & 999 \\
it-es & - & - & - & 6,614 & 877 & 885 & - & 929 & 999 \\
pt-en & 9,158 & 3,590 & 4,254 & 4,918 & 1,747 & 2,286 & 29,940 & 1,002 & 1,019 \\
pt-es & - & - & - & 3,132 & 1,218 & 1,256 & - & 1,001 & 1,018 \\
ru-en & 12,112 & 9,497 & 8,634 & - & - & - & 4,829 & 970 & 1,124 \\ \hline
\end{tabular}
\end{table*}

\end{document}